\documentclass[sigconf]{acmart}

\AtBeginDocument{%
  \providecommand\BibTeX{{%
    \normalfont B\kern-0.5em{\scshape i\kern-0.25em b}\kern-0.8em\TeX}}}

\copyrightyear{2024}
\acmYear{2024}
\setcopyright{none}
\acmConference[ETRA '24]{2024 Symposium on Eye Tracking Research and Applications}{June 4--7, 2024}{Glasgow, United Kingdom}
\acmBooktitle{2024 Symposium on Eye Tracking Research and Applications (ETRA '24), June 4--7, 2024, Glasgow, United Kingdom}

\acmISBN{}
\usepackage{wrapfig}
\usepackage{xspace}
\usepackage{subcaption}
\usepackage{hyperref}

\usepackage{array}

\newcolumntype{P}[1]{>{\centering\arraybackslash}p{#1}} 
\newcolumntype{R}[1]{>{\raggedleft\arraybackslash}p{#1}} 

\begin{document}


\title{Swap It Like Its Hot: Segmentation-based spoof attacks on eye-tracking images}

\author{Anish S. Narkar}
\email{anishnarkar@vt.edu}
\affiliation{%
  \institution{Virginia Tech}
  \city{Blacksburg}
  \state{Virginia}
  \country{USA}
  \postcode{24060}
}

\author{Brendan David-John}
\email{bmdj@vt.edu}
\affiliation{%
  \institution{Virginia Tech}
  \city{Blacksburg}
  \state{Virginia}
  \country{USA}
  \postcode{24060}
}

\renewcommand{\shortauthors}{Narkar and David-John}

\newcommand{\name}{\textsc{IrisSwap}\xspace}

\begin{abstract}

Video-based eye trackers capture the iris biometric and enable authentication to secure user identity. However, biometric authentication is susceptible to spoofing another user’s identity through physical or digital manipulation. The current standard to identify physical spoofing attacks on eye-tracking sensors uses liveness detection. Liveness detection classifies gaze data as real or fake, which is sufficient to detect physical presentation attacks. However, such defenses cannot detect a spoofing attack when real eye image inputs are digitally manipulated to swap the iris pattern of another person. We propose \name as a novel attack on gaze-based liveness detection. \name allows attackers to segment and digitally swap in a victim’s iris pattern to fool iris authentication. Both offline and online attacks produce gaze data that deceives the current state-of-the-art defense models at rates up to 58\% and motivates the need to develop more advanced authentication methods for eye trackers.

\end{abstract}





\maketitle

\setlength{\belowcaptionskip}{0pt}
\setlength{\abovecaptionskip}{2pt}
\setlength{\textfloatsep}{12pt}

\section{Introduction}

Presenting falsified biometric samples of authentic users to access a system is referred to as a \textit{presentation} attack. Presentation or spoofing attacks can be performed using biometric traits such as fingerprints, faces, and irises. 
Iris attacks primarily present physical spoofs using contact lenses~\cite{Venkatesh2019ANM}, printouts~\cite{Ref2,Ref3} and playing back recorded videos~\cite{Ref5}. Spoofing attacks can lead to significant harm to individuals. For example, facial authentication for phones and laptops enables adversaries to access sensitive information, including bank accounts~\cite{jarrahi_2021}. The significance of such harms is clear within intimate partner violence, in which an adversarial but authenticated user can monitor activity, impersonate a victim, or install spyware~\cite{freed2018stalker,chatterjee2018spyware}. 
 While face biometrics are more common on mobile devices today, iris biometrics are seeing increased use in mixed-reality where unobstructed face images are not available. The Microsoft HoloLens2, Magic Leap 2, and Apple Vision Pro enable iris authentication through their integrated eye-tracking sensors~\cite{hololens2,magic_leap_iris,vision_pro_iris}.

Current iris spoof detection methods use gaze estimation to detect the presentation of a physical spoof~\cite{czajka2018presentation}. The state-of-the-art defense method extracts gaze velocity signals for \textit{liveness detection}, i.e., whether the eye tracking inputs are from a real eye or a physical spoof~\cite{Raju}. 
To our knowledge, a real-time iris presentation attack that digitally manipulates real eye images to beat liveness detection has not been demonstrated. In this paper, we propose and demonstrate that \name can be used to deceive gaze-based liveness detection models and successfully spoof the iris biometric. Our primary contributions are a pipeline for digital spoof manipulations in eye images\,(Sec.\,\ref{sec:sys_design}) and an evaluation of the attack against state-of-the-art liveness detection in both offline\,(58\% user-level attack success rate) and online\,(55\% user-level attack success rate) eye-tracking pipelines\,(Sec.\,\ref{sec:pipeline_result}).



\begin{figure*}[h]
  \centering
  \includegraphics[width=\linewidth]{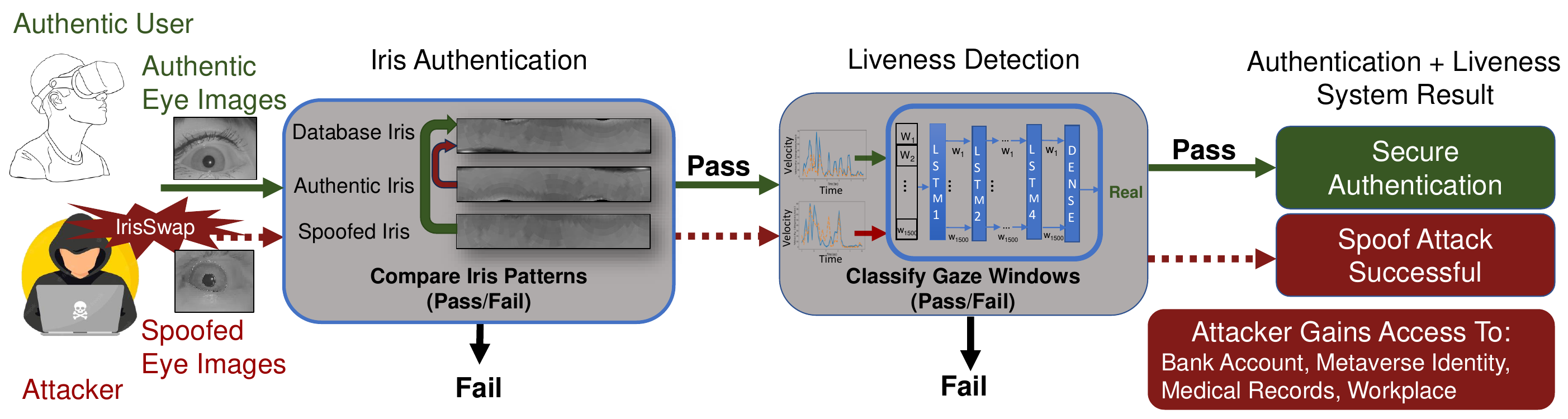}
  \caption{Illustration of the spoofing attack pipeline for iris patterns with gaze-based liveness detection.}
  \label{fig:design}
\end{figure*}
\section{Background and Motivation}

Biometric authentication is a secure method to prevent unauthorized access to systems due to their universality, distinctiveness, permanence, and collectability~\cite{jain2004introduction}. 
Liveness detection is applied to determine if the presented biometric credentials originated from an authentic source. For face images, depth sensors on mobile phones are used to determine if the presented face is a 3D face or a 2D printout before authenticating the user~\cite{facedefense}. %
For eye images containing the iris, which is a gold standard biometric, physical presentation attacks include contact lenses~\cite{Venkatesh2019ANM}, printouts~\cite{Ref2,Ref3} and displaying videos~\cite{Ref5}. Textured contact lenses and high-quality printouts require expertise and specific equipment. This requirement limits the breadth of such attacks and lowers the impact of the threat posed by such attacks. For contact lenses, techniques that process the sensor image to identify features introduced by contacts are effective in detecting textured spoof attacks~\cite{doyle2015robust}, while video playback attacks can be detected through Eulerian video magnification with as low as eleven video frames~\cite{Ref5}. A digital manipulation attack does not require special equipment and can be performed if you only have the iris pattern of the victim. 

Figure~\ref{fig:design} demonstrates a typical iris authentication and liveness detection mechanism. The attacker presents spoofed eye images (generated using contact lenses, printouts, physical video playbacks, or digital manipulation) to the system. The system first performs the iris authentication and then performs a liveness detection. Access is granted only if both these checks are satisfied.

The standard liveness detection defense for eye images performs gaze estimation and classifies real gaze signals based on micro-saccades and fixations~\cite{rigasgaze}, ocular plant characteristics~\cite{oculartplant} and velocity profiles~\cite{em1,Raju}. 
These defense mechanisms are successful against impersonation using static presentations such as eye patches. Spoof eye patches cover most of the attacker's eye and strongly impacts the gaze estimation signal enabling detection by liveness models trained on gaze velocity. However, a digitally manipulated image would replace the iris with the victim pattern without restricting their eye movements. Thus, the success of digital manipulation attacks is based on the effectiveness of the iris swapping method.

Raju et al.~\cite{Raju} present the latest work on eye-tracking liveness detection using gaze velocity signals, as illustrated in Figure\,\ref{fig:design}. The authors trained a machine-learning model on horizontal and vertical components of gaze velocity to determine real or spoof. The authors evaluated against physical spoofing attacks in the form of an iris pattern printed on paper and worn on an eye patch. The proposed method reliably detected spoof attacks from the eye patch with an accuracy of 98\%. 
The liveness detection approach reliably detected attacks, as the eye patch produced a gaze signal that was clearly not from a real eye. 
Thus, an attack with the ability to spoof an iris pattern without impacting gaze estimation poses a risk for current liveness detection-based defenses.

Chaudhary et al.~\cite{rubbersheetmodel} demonstrated high-accuracy swapping of the iris pattern using deep segmentation networks. Their work presented a privacy solution by removing the user's iris pattern from recorded eye-tracking data. In contrast, our work instead overlays the iris pattern of a victim to perform a presentation attack. Through accurate and efficient swapping of the iris pattern, authentication can be spoofed without impacting the gaze signal. 

\section{Attack Model And Methodology}
The \name attack has several assumptions and necessary definitions. The \textit{challenge} refers to the task that will be analyzed to determine the liveness of the input data~\cite{Face}. The \textit{attacker} is the person who is carrying out the attack and the \textit{victim} is the person whose iris image will be spoofed by the attacker. We assume that the attacker possesses an iris image of the victim. The attacker applies the spoof to a recording of their own eye-tracking images as they perform the challenge task for \textit{offline attacks} or make use of software plugins to manipulate their eye image stream in real-time for \textit{online attacks}. The attacker is successful if their sequence of eye images pass both the liveness detection classification and iris authentication as the victim. 
We set our assumptions for data access based on current standards for liveness detection and iris authentication, meaning the spoof detection model is limited to processing only gaze estimates produced by the manipulated inputs while the authentication module is limited to the eye image with the manipulation applied. While a spoof detection approach could observe the modified eye images to detect manipulations, that is outside the scope of our work and requires the design of a novel defense method and evaluation. 

\subsection{Design Challenges}
\subsubsection{Challenge 1: Accurate segmentation and efficient swapping}
A successful attack will accurately segment and replace the attacker's iris pattern with the victim's iris. The swap should take into account any physical differences between eye images, i.e., pupil diameter and eye orientation. For online attacks, the processing pipeline must execute fast enough to support gaze estimation during the challenge task. 

\subsubsection{Challenge 2: Minimal impact on gaze estimation} A successful attack will not cause significant deviation between the gaze estimates of swapped and unmodified images as they are used for liveness classification. Accuracy and precision of gaze data are computed to measure the difference between swapped and unmodified inputs and liveness detection models are re-trained on for each experiment to evaluate whether the manipulation is detected within the gaze signal. 

\subsection{\name System Design}
\label{sec:sys_design}

\begin{figure*}[h]
  \centering
  \includegraphics[width=\linewidth]{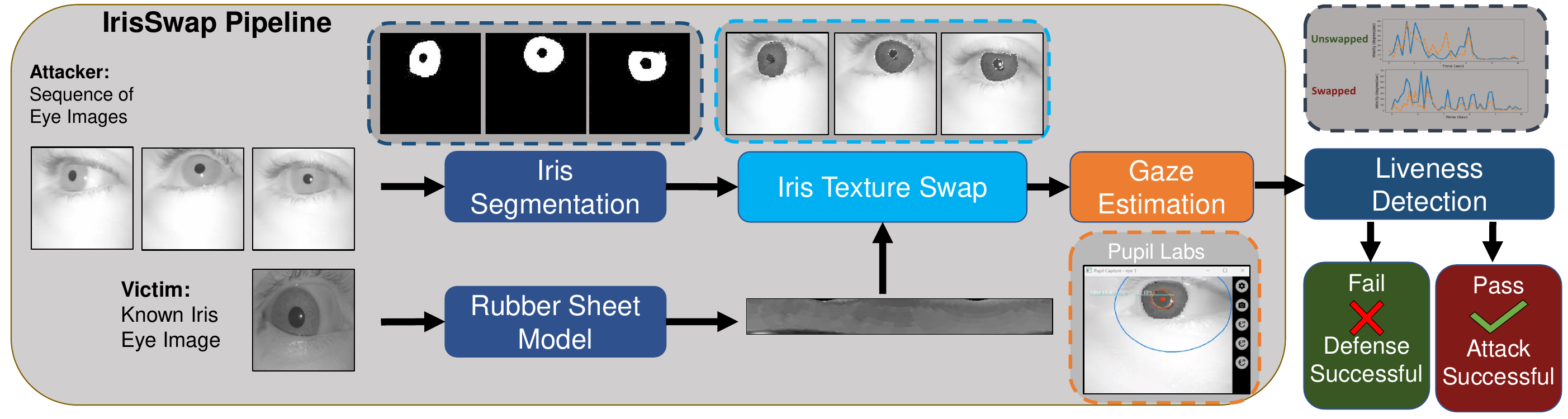}
  \caption{\name pipeline showing the flow of data through iris segmentation and swapping before gaze estimation is applied. The output gaze positions are processed into windows of gaze velocity that are classified by a liveness detection model.}
  \label{fig:system}
\end{figure*}

Inputs to our pipeline are a sequence of eye images from the attacker and the iris pattern of a victim\,(Fig.\,\ref{fig:system}). \name replaces the iris pattern of the attacker with that of the victim and generates gaze positions from the modified eye images. Each stage of the pipeline requires independent models for data processing. 


\subsubsection{Iris Segmentation}
\label{sec:method_seg}
Iris segmentation is applied to raw grayscale images of the eye. We adopted the RITNet segmentation model based on high performance in the rubber sheet iris protection work~\cite{rubbersheetmodel}. 
However, when testing with RITNet we observed that segmented images contained stray iris class pixels outside of the iris region. We instead considered a double UNET architecture to obtain finer and more accurate segmentations~\cite{doubleunet}. 
RITNet is based on a single UNET model, and it was not feasible to duplicate RITNet twice given the size of the RITNet multi-stage pipeline. Thus, we implemented a new segmentation model based on the double UNET architecture called Shallow-net. 
Please see the Supplementary Material for a complete description of the Shallow-net architecture. 


\subsubsection{Iris Texture Swap}

Iris swapping uses the segmentation mask from the attacker's eye image to replace the iris with that of the victim to spoof their identity. The inverse rubber sheet model transforms an iris pattern in polar coordinates and accounts for varying shape and orientation by leveraging inner and outer radius of the target segmented region~\cite{rubbersheetmodel}. 
A more detailed description of how this process is implemented can be found in the original paper~\cite{rubbersheetmodel}.

\subsubsection{Gaze Estimation}
Gaze positions are estimated using the Pupil Labs hardware and software suite\,(v3.5.1) and the standard four-point validation procedure to compute gaze accuracy and precision metrics~\cite{kassner2014pupil}. We used the Pupil Labs 3D landmark-based gaze model with default parameters to generate gaze positions which are then processed into horizontal and vertical velocity components for liveness detection.   

\subsubsection{Challenge Task} The challenge task in Raju et al.~\cite{Raju} had users observe a single static point on a screen. 
We introduced a challenge task with multiple targets to increase the difficulty in passing liveness detection using velocity-based models. 
The Pupil Labs calibration and validation routines were used as the challenge tasks they feature five and four target positions, respectively. Furthermore, using the calibration step for the challenge allows liveness detection to be organically integrated into the eye tracker setup without an additional step. 

\subsubsection{System Deployment}
\label{sec:method_deployment}
Python code for segmentation and swapping was implemented within the Pupil Player for offline attacks on previously recorded data and was implemented in Pupil Capture as a plug-in to apply online attacks.

\section{Datasets and Experiments}
To evaluate the \name pipeline we conducted experiments on offline and online attacks. Implementing the pipeline required training an iris segmentation model for Pupil Labs eye images and training a liveness detection classifier based on Pupil Labs gaze signals. 
In Section~\ref{sec:results_dataset}, we define the challenge task and dataset for our experiments. 
In Section~\ref{sec:pipeline_test}, we describe the optimization and performance for each component of \name. In Section~\ref{sec:metrics}, we define the metrics to evaluate \name in terms of gaze estimation and attack success rate, then present our evaluation results in Section~\ref{sec:pipeline_result}.
    
\subsection{Datasets}
\label{sec:results_dataset}

\paragraph{Offline Attack Dataset}
For offline analysis, we used a dataset of eye-tracking recordings previously evaluated in the context of iris authentication provided by John et al.~\cite{john2020security}. This dataset contains data from 15 subjects performing a five-point calibration task and validation using the Pupil Labs Pro\,(2016) eye tracker at 30Hz. We chose this dataset as the data aligned with the challenge task and could be re-analyzed with the Pupil Labs software. Subjects sat in front of a computer monitor and were presented with one calibration point at a time for four seconds. 

\paragraph{Online Attack Dataset}
To evaluate the real-time performance of \name we collected a new dataset using a Pupil Labs Core\,(2022) eye tracker with an IRB-approved user study with 18 subjects (13M, 5F, Avg. Age = 27.2 $\pm$ 1.2 years). Subjects sat approximately one meter in front of a computer monitor and were presented with the challenge task\,(Sec.\,\ref{sec:challenge}), with each target shown for four to six seconds. The study consisted of two conditions, swapped and unswapped based on whether \name was enabled or not and were counterbalanced across subjects.


\paragraph{Iris Segmentation}
We used 2600 ground truth images from CASIA~\cite{casiaseg} to train our initial iris segmentation model\,(Sec.\,\ref{sec:method_seg}). We then isolated data from two subjects of the offline dataset to fine-tune a model for offline analysis. Data from these two subjects were withheld from further evaluations against liveness detection. We also performed a pilot study to test the online attack using the Pupil Labs Core eye tracker. We labeled the iris images from this data to fine-tune a separate segmentation model for the online attacks. This ensured that the segmentation results were consistent for the iris sequences from each Pupil Labs eye tracker. We manually segmented a total of 1000 eye images from both the offline\,(750) and pilot dataset\,(250) for fine-tuning. For fine-tuning each model, 70\% of the images were used for training, 15\% for validation, and the remaining 15\% images were used to evaluate model performance. 

\paragraph{Liveness Detection} \label{sec:challenge}
To train and evaluate the liveness detection models we assigned subjects to either the training or the test set. The training set consisted of data from 60\% of the subjects\,(offline: 7, online: 11) and the test set consisted of the remaining 40\% subjects\,(offline: 6, online: 7). While training we further divided the training set into a validation set. The validation set consisted of 30\% of randomly selected subjects from the training set. 
The instantaneous horizontal and vertical velocity signals from each user were computed based on the gaze estimates produced by Pupil Labs. Velocity components over 800 $^\circ/sec$ were removed and replaced via interpolation~\cite{dowiasch2015effects}. Min-max normalization between zero and one was applied to the velocity signals for each user to avoid any data leakage.  

\subsection{\name Component Testing and Metrics}
\label{sec:pipeline_test}
\paragraph{Iris Segmentation}
As described in \textbf{Challenge 1}, the effectiveness of \name relies on accurate and efficient iris segmentation. We compared segmentation results using RitNet and our Shallow-net model using the Dice score metric~\cite{Carass2020}. The Dice score ranges from zero to one and measures the degree of overlap between predicted and ground truth while accounting for class imbalance and is analogous to the F1 score for segmentation tasks. 
The computed Dice score for RITNet on the testing set was 0.91, while the Dice score for Shallow-net was 0.96. 
Based on these results, we used Shallow-net to evaluate and deploy \name.

\paragraph{Liveness Detection}
A successful \name attack is determined by its ability to swap irises without producing gaze estimates flagged as fake by the liveness model. 
The liveness detection model used by Raju et al~\cite{Raju} was based on a customized ResNet and was developed using a dataset that was collected on a 1000Hz device. 
Due to the lower sampling rate in our data, it was not possible to use their exact architecture or model weights. A pilot study with the online \name attack identified a drop in sampling rate as low as 3Hz. To avoid issues with oversampling online gaze samples during deployment, we decided to downsample all data to 3Hz for the liveness detection. 
We optimized their proposed architecture on our data and compared performance with LSTM models following their training/testing protocol using window length and step size as the hyperparameters. The full description of this optimization and comparison is described in the Supplementary Material. We found our LSTM models with a window length of seven samples\,(2.33 seconds) and step size of three performed best with an attack classification rate of 99.9\% and deployed them for analysis. 



\paragraph{Iris Authentication}
A successful \name attack must pass a standard iris authentication for the victim's iris. For offline analysis, subject S013 was randomly chosen as the victim of the attack. For the online attack, we needed to assign the victim iris before the study due to the real-time nature of the attack. An iris pattern from one of the authors using the Pupil Labs Core was considered the victim of the online attack. The data from these subjects were excluded for training and development of all the components of the \name pipeline and the liveness detection model. The standard metric for iris authentication is Hamming Distance\,(HD) and is computed between two iris biometric templates~\cite{daugman}. HD measures the number of bits that are different between the biometric templates. The phase-based Daugman method using 2D Gabor wavelets was used for extracting binary iris templates. A lower HD indicates a closer match, with HD less than 0.37 being considered a sufficient condition for genuine authentication from eye tracker images~\cite{john2020security}. We selected ten random frames from each of the spoofed output sequences and compared them with the iris template from the corresponding frame number of the victim's eye-tracking sequence. 

\subsection{Metrics}
\label{sec:metrics}
\paragraph{Gaze Estimation} 
As described in \textbf{Challenge 2}, \name should not have a significant effect on the accuracy and precision of the gaze signal. Accuracy is the average deviation between the gaze positions 
obtained by the eye-tracker and the location of the validation targets. Precision is the Root Mean Squared angular distance between successive samples. Both these metrics are computed on the validation targets of the challenge task using the Pupil Player software. 

\paragraph{Attack Success Rate}

Attack Success Rate\,(ASR) is used to measure the performance of our attacks. ASR measures the proportion of spoofed samples that were misidentified by the liveness models. ASR for a perfect attack would have a value of one, while for a perfect defense, it would be zero. 
To calculate ASR we first feed the velocity windows to the liveness detection model. ASR is calculated on two levels. First, at the window level, ASR is calculated based on individual window predictions for all the users. Second, on the user level ASR was calculated based on model prediction for the entire user sample data. The model predictions of the majority of window samples for each user is assigned as the model prediction for that user. The ASR metrics are defined as 

$$ ASR_{window} = \frac{Spoofed\,\, windows\,\, classified\,\, as\,\, Real}{Total\,\, \#\,\, of\,\,\, Spoofed\, windows}$$ \\ $$ASR_{user} = \frac{Spoofed\,\, users\,\, classified\,\, as\,\, Real}{Total\,\,\#\,\, of\,\,\,users}.$$

The reliability of attack success depends on the random train-test split of the subjects. We generated ten random train-test splits to train the liveness detection model and evaluate the resulting variance. 
The reported mean ASR across all splits is used to demonstrate the effectiveness of the \name pipeline. 

\subsection{Attack Results}


\label{sec:pipeline_result}
We evaluated the offline and online impact of \name on gaze estimation, computed the ASR against an optimal liveness detection model, and confirmed that spoofed eye images pass iris authentication for the victim.

\subsubsection{\name Impact on Iris Authentication}
\label{sec:iris_auth_impact}
The rightmost column of Table~\ref{tab:accuracy} shows the result of the iris authentication of the selected victim. The average HD of the spoofed samples for offline and online scenarios was 0.36 and 0.39, respectively. The HD values for each subject in the offline and online evaluations are listed in the Supplementary.  
The mean HD for offline attacks was marginally under the 0.37 threshold which is a commonly accepted limit and aligns with past eye-tracker authentication studies~\cite{daugman,john2020security,eyeveildataset}. The victim was authenticated for ten out of the thirteen subjects in offline attacks. The mean HD for online attacks was 0.39 which was marginally above the threshold; however, the victim was authenticated for thirteen out of the eighteen subjects. Outlier values of HD\,(0.45, 0.53, and 0.57) for three of the subjects increased the overall mean HD for online attacks. Thus, for offline and online attacks the victim identity was successfully spoofed for 77\% and 72\% of the users, respectively.

\begin{figure*}[h]
    \centering
    \includegraphics[width=0.9\linewidth]{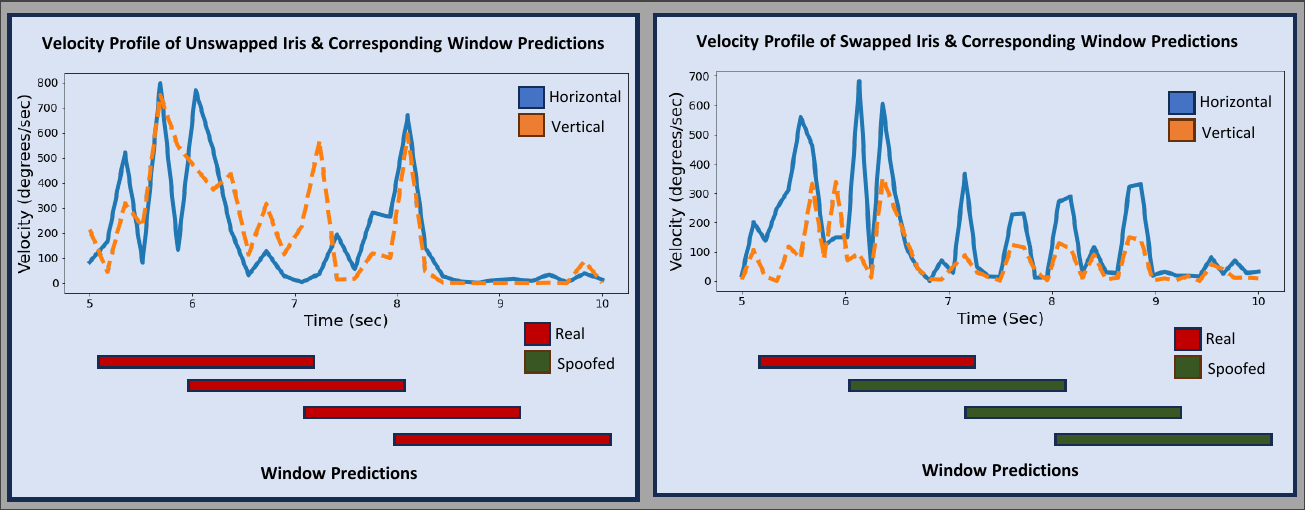}
  \caption{Velocity profiles for Unswapped (Left) and Swapped (Right) samples over five seconds of an online attack. A similar profile is produced between 5s and 7s while the differences between 7s and 9s are flagged as spoofs by the liveness model.}
  \label{fig:velocity}
\end{figure*}


\subsubsection{Eye-Tracking Data Quality}
Figure\,\ref{fig:velocity} visualizes the impact of \name on the velocity signals extracted from the gaze estimates, and Table~\ref{tab:accuracy} presents the impact on gaze estimation metrics. The difference between the mean accuracy of spoofed and unmodified data for offline and online attacks are 0.83$^\circ$ and 1.7$^\circ$, respectively. For the offline attacks sampling rate was preserved, and manipulations impacted the pupil landmark tracking used for gaze estimation producing less than a degree of error in terms of accuracy. 
For the online dataset, the sampling rate was reduced by a factor of 8.9 $\pm$ 2.6 on average when \name was enabled compared to the unswapped condition. The lowest sampling rate produced was 3Hz. The reduced sampling rate resulted in fewer samples for Pupil Labs to perform an accurate calibration and for computing the validation metrics; however, it still provided enough data to enable the Pupil Labs calibration and our challenge task. Despite the introduced spatial error, the gaze velocity peaked in similar regions (between 5 and 7 seconds) as shown in Figure~\ref{fig:velocity}. However, an impact from the reduced sampling rate can be observed between 7 and 9 seconds. These differences in the velocity profile are detected by the liveness models. 

\subsubsection{\name Effectiveness}
\label{sec:attack_effectiveness}
\begin{figure}[H]
    \centering
    \includegraphics[width=\linewidth]{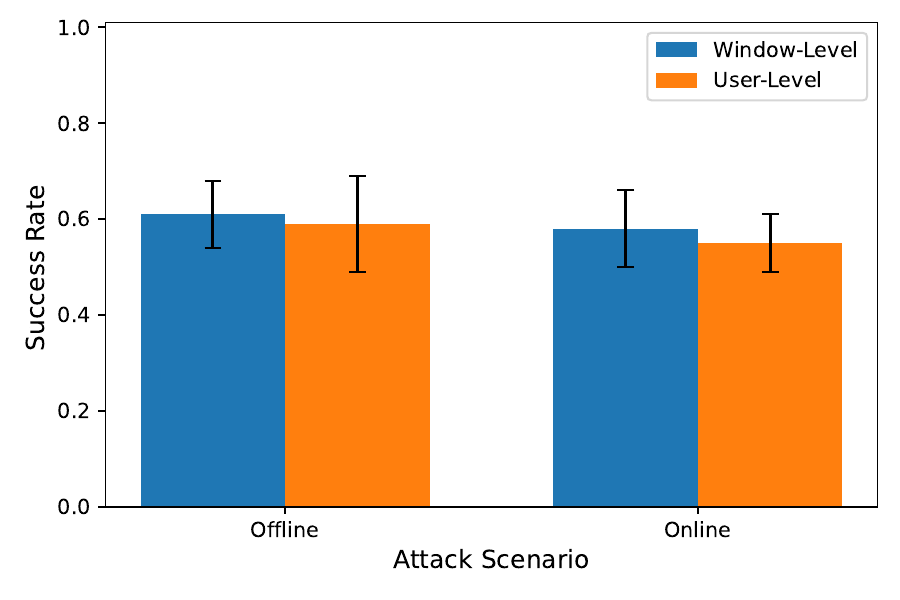}
    \caption{Window-level ASR indicates the success rate calculated based on each window of the test-user data. Window-level ASR is 0.61 $\pm$ 0.07 and 0.58 $\pm$ 0.08 for offline and online attacks, respectively. User-level ASR makes one real or spoof classification based on all windows from a single user. User-level ASR is 0.59 $\pm$ 0.10 and 0.55 $\pm$ 0.06 for offline and online attacks, respectively.}
    \label{fig:results_qualitative}
\end{figure}

Figure~\ref{fig:results_qualitative} shows the attack effectiveness of the \name pipeline. The average $ASR_{window}$ for offline and online scenarios is 0.61 and 0.58, respectively. The average $ASR_{user}$ for offline and online attacks is 0.59 and 0.55, respectively. These results indicate that the \name pipeline is marginally more successful in offline scenarios and in both cases just under half the attacks were detected by liveness detection. Based on the increase in gaze error and visualization of velocity profiles, we conclude that ASRs below 100\% can be attributed to lower sampling frequencies and the image manipulation degrading gaze estimation accuracy. Still, an ASR above 50\% poses a risk if attacks were mounted on a large scale.


\begin{table*}[]
\caption{Impact of \name on eye-tracking data quality across subjects for online and offline subjects. \name only introduced 0.83$^\circ$ of gaze error on average for offline condition but introduced a gaze error of 1.7$^\circ$ for real-time application. The rightmost column shows the mean HD for all the train-test splits. 
}
\label{tab:accuracy}
\resizebox{\textwidth}{!}{%
\begin{tabular}{|l|cc|cc|c|}
\hline
Application & \begin{tabular}[c]{@{}c@{}}Unswapped  Accuracy \\ (degrees)\end{tabular} & \begin{tabular}[c]{@{}c@{}}Swapped Accuracy \\ (degrees)\end{tabular} & \begin{tabular}[c]{@{}c@{}}Unswapped Precision\\ (degrees)\end{tabular} & \begin{tabular}[c]{@{}c@{}}Swapped Precision\\ (degrees)\end{tabular} & \multicolumn{1}{l|}{Hamming Distance} \\ \hline
Offline & 0.56 $\pm$ 0.33 & 1.39 $\pm$ 0.40 & 0.12 
 $\pm$ 0.03 & 0.12 $\pm$ 0.03 & 0.36 $\pm$ 0.02 \\
Real-time & 1.91 $\pm$ 0.65 & 3.56 $\pm$ 0.48 & 0.13 $\pm$ 0.05 & 0.25 $\pm$ 0.05 & 0.39 $\pm$ 0.06 \\ \hline
\end{tabular}%
}
\end{table*}

\section{Conclusion and Discussion}
    Current state-of-the-art defenses against iris presentation attacks employ a gaze velocity-based approach. Our \name pipeline digitally manipulates an attacker's iris to match that of a victim while enabling gaze estimation. Our method successfully attacked a state-of-the-art liveness detection model architecture at a rate of 59\% and 55\% at a user level in offline and online attacks, respectively. Our attack shows that the current standard for gaze-based liveness detection cannot reliably detect presentation attacks produced by digital manipulations. The \name pipeline demonstrates a new class of digital presentation attacks that differ from the playback attacks used in prior works. The real-time nature of \name and a success rate over 50\% indicates a practical security concern if such an attack is optimized and mounted on a large scale.

\paragraph{Implications}
The proposed \name attack is the first system using digital manipulation to spoof iris-based authentication in an eye-tracking system successfully. Our attacks were successful 58\% of the time, meaning they were neither perfect nor benign. The threat we identified can still be mitigated 
as we expect future solutions in the field to detect \name and similar digital manipulations. For example, swapped eye images may have spurious white pixels or the victim iris pattern may have clear differences in grayscale intensity from the original eye\,(see Figure\,\ref{fig:system}, Iris Texture Swap). A defense model that processes the swapped eye images could detect such artifacts. Our results also suggest that one of the reasons an attack is not successful is that the reduced sampling rate affects the accuracy of calibration and the consistency of velocity profiles leading to spoof detection. If our attack code was integrated in a more optimal manner, and not through a patch-like approach using Python plug-ins at runtime, then we expect the attack to have a higher success rate.

\paragraph{Limitations}
Our liveness detection evaluation considered eye movement data from a head-worn eye tracker downsampled to 3Hz, as opposed to an EyeLink desktop tracker at 1000Hz in existing work on liveness detection~\cite{Raju}. Eye-tracking systems with more stable calibration models or higher sampling rates may produce different results in terms of successfully detecting attacks. Our findings on HD and iris authentication depend on infrared images captured by the Pupil Labs system at 320 $\times$ 240. While this resolution is sufficient for authentication~\cite{eyeveildataset,phillips2011impact}, higher-resolution images would make for more convincing spoof attacks. Our segmentation model relied on the binary classification of which pixels were the iris region, though a multi-class approach may be more accurate and allow for a smoother application of the inverse rubber sheet.


\paragraph{Future Work}
First, we plan to implement and evaluate a suite of defense measures that could include embedding the output of the eye camera with a digital watermark~\cite{plata2020robust}, training a classifier to detect visual artifacts resulting from \name, or the integration of peri-ocular biometrics that feature eye shape and brows~\cite{woodard2010fusion}. Second, now that there is proof that iris swaps can beat liveness detection, we expect to further study new attack models. For example, GANs or Variational Autoencoders commonly used in deep fakes can be integrated to generate high-resolution eye images that retain gaze direction with a swapped iris in one step, without relying on a specific segmentation model. Balancing online performance with naturalistic-looking spoofs presents a key research challenge in this context to benchmark how effective an iris spoofing attack can become.  

\paragraph{Privacy and Ethics}
Privacy and ethics considerations are critical as we identify a new security vulnerability in iris authentication systems that use gaze estimation for liveness detection. First, given the scope of a short paper, we have identified and presented results on the feasibility of the attack, but do not present an optimized system or evaluate a corresponding defense mechanism. We noted potential defense mechanisms for the current attack and expected extensions of the system in our discussion. Second, we took great care to protect the privacy of our study subjects who were necessary to enable our experiments. Within the manuscript, we ensured that all eye images used to make figures were outside the quality standards for iris biometrics outlined in ISO/IEC 19794-6:2005~\cite{formats20056}, including purposely saving images in a compressed JPEG format and ensuring the image was first resized such that the iris diameter spanned less than 200 pixels. The Supplementary video during the attack sequence was heavily compressed as well. Our IRB-approved study provided informed consent on the purpose of the study, the critical nature of securing biometrics, and our data management approach. Our collected research dataset will not be posted publicly but gaze sample data may be shared with validated research teams in which data usage will be moderated by the authors.

\bibliographystyle{ACM-Reference-Format}
\bibliography{sample-base}

\appendix

\end{document}